\documentclass[runningheads]{llncs}
\usepackage[T1]{fontenc}
\usepackage{graphicx,verbatim}
\usepackage{amssymb}
\usepackage{amsmath}
\usepackage{bm}
\usepackage{comment}
\usepackage{amsmath,amssymb,amsfonts}
\usepackage{algorithmic}
\usepackage{graphicx}
\usepackage{textcomp}
\usepackage{multirow}
\usepackage{booktabs}
\usepackage{hyperref}
\usepackage{tabularx}
\usepackage{xcolor}
\usepackage{multirow}
\usepackage{booktabs}
\usepackage{bbding}        
\usepackage{hyperref}
\usepackage{color}

\begin{document}
\title{Spatially Gene Expression Prediction using Dual-Scale Contrastive Learning}

\author{Mingcheng Qu\inst{1} \and Yuncong Wu\inst{2} \and Donglin Di\inst{3} \and Yue Gao\inst{3} \and Tonghua Su\inst{1} \and Yang Song\inst{4} \and Lei Fan\inst{4}\Envelope}
\authorrunning{Qu M. et al.}
\institute{
   Faculty of Computing, Harbin Institute of Technology, Harbin, China \and
    School of Astronautics, Harbin Institute of Technology, Harbin, China \and
    School of Software, Tsinghua University, Beijing, China \and
    School of Computer Science and Engineering, UNSW, Sydney, Australia\\
    \email{lei.fan1@unsw.edu.au}}
    
\maketitle              

\begingroup                                       
  \renewcommand\thefootnote{\Envelope}
  \footnotetext{Corresponding author}
\endgroup

\begin{abstract}
Spatial transcriptomics (ST) provides crucial insights into tissue micro-environments, but is limited to its high cost and complexity. As an alternative, predicting gene expression from pathology whole slide images (WSI) is gaining increasing attention. However, existing methods typically rely on single patches or a single pathology modality, neglecting the complex spatial and molecular interactions between target and neighboring information (\textit{e.g.}, gene co-expression). This leads to a failure in establishing connections among adjacent regions and capturing intricate cross-modal relationships.
To address these issues, we propose NH²2ST, a framework that integrates spatial context and both pathology and gene modalities for gene expression prediction. Our model comprises a query branch and a neighbor branch to process paired target patch and gene data and their neighboring regions, where cross-attention and contrastive learning are employed to capture intrinsic associations and ensure alignments between pathology and gene expression.
Extensive experiments on six datasets demonstrate that our model consistently outperforms existing methods, achieving over 20\% in PCC metrics. 
Codes are available at \url{https://github.com/MCPathology/NH2ST}.


\end{abstract}

\section{Introduction}

Gene expression plays a crucial role in understanding cellular functions and disease mechanisms. 
Recently, spatial transcriptomics (ST) \cite{ST-intro1} has emerged to integrate both spatial coordinates and transcriptomic signatures \cite{transcriptomic}, enabling spatially resolved gene expression profiling \cite{ST}. Despite its advantages, the high cost and technical complexity limit its scalability \cite{ST-limitation}. This highlights the need for computational approaches capable of predicting spatially resolved gene expression directly from whole slide images (WSIs) \cite{ST-need-DL,fan2022cancer,fan2021learning}.

Earlier studies \cite{ST-Net,patch-predict} treated individual patches as input with gene expression as labels. Some approaches then incorporated spatial context, such as HisToGene \cite{HistToGene} and mclSTExp \cite{mclSTExp}. TRIPLEX \cite{TRIPLEX} integrated features from the entire neighborhood patch. Hist2ST \cite{Hist2ST} further improved upon this by using graphs to capture neighborhood relationships. 
However, these methods primarily focus on capturing features from a broader context coarsely (\textit{e.g.}, extracting features from larger pathology patches) without explicitly modeling inter-patch relationships, limiting their ability to fully capture the complex spatial and molecular interactions present in pathology images \cite{Hist2ST,qu2024boundary}. Since ST data inherently displays strong spatial continuity \cite{ST-tissue}, and gene expression in adjacent locations is often correlated due to biological factors like cell-cell interactions \cite{cell-interactions} and micro-environment \cite{HistToGene,tang2025prototype}. Consequently, leveraging the spatial context from neighboring patches and establishing their interactions are essential.

On the other hand, most previous methods \cite{ST-Net,HistToGene,Hist2ST} rely on only pathology inputs for predicting the gene expression. Some studies \cite{EGN} explored example-based learning to retrieve relevant samples, while recent works \cite{BLEEP,mclSTExp} employed contrastive learning \cite{CL,fan2022fast} to align these pathology and gene expression modalities.
However, while contrastive learning enforces alignment between pathology image features and gene expression features at the same spatial location, it does not fully consider the interactions between the two modalities. Pathology features are often associated with specific gene expressions (\textit{e.g.}, invasive margins strongly correlate with PD-L1 expression \cite{PD-L1}), requiring a more dynamic and integrative approach to aggregate information from both modalities \cite{qu2025multimodal}.

To address these challenges, we aim to integrate spatial neighborhood information and multimodal interactions between pathology and gene expression, where hypergraph learning \cite{hypergraph,jing2025multi} is employed to capture neighboring relationships, while cross-attention \cite{CA} and contrastive learning \cite{CL,jing2025multi} are used to model interactions and ensure alignment between the two modalities. 
In this paper, we propose NH²2ST, Neighbor-guided Hypergraph Histopathology to ST gene expression. It consists of two branches: 1) A query branch processes the paired target patch and ST data, where cross-attention and cross-modal contrastive learning are applied to refine and align pathology-gene features. 2) A neighbor branch extends this by incorporating the target patch along with its neighboring regions and paired ST data, leveraging hypergraphs to model interactions among different patches and their ST spots. The neighbor branch plays an auxiliary role, supporting the query branch in training more robust feature extractors. Meanwhile, the pathology extractor in the query branch is further enhanced with a prediction translator to generate gene expression values during inference.

Our contributions can be summarized as: 
We propose a dual-scale framework for predicting spatially resolved gene expression from pathology patches, employing cross-attention and contrastive learning to enhance alignment between the pathology and gene modalities. 
We leverage hypergraph learning to model the interactions among different patches for both modalities, capturing complex spatial and molecular interactions effectively. 
Extensive experiments on six publicly available datasets demonstrate that our model outperforms advanced methods.  
\section{Method}

\subsection{Overview}
\textbf{Preliminaries.} Given a WSI $X$ represented as \(M\) non-overlapping tissue patches, \( \{x_i\}_{i=1}^{M}\), where each patch \(x_i \in \mathbb{R}^{h \times w \times 3}\) corresponds to a ST spot detected by a probe, paired with a gene expression vector \(y_i \in \mathbb{R}^n\) capturing the normalized expression levels of \(n\) genes. To mitigate technical noise and enhance signal quality, we followed previous studies \cite{ST-Net,TRIPLEX} by selecting \(n\)=250 most highly expressed genes, applying a logarithmic transformation to their expression values. The objective is to learn a mapping function \(\phi: x_i \rightarrow y_i\) to predict gene expression from corresponding patches.

\begin{figure*}[t]
    \centering
    \includegraphics[width=1\textwidth]{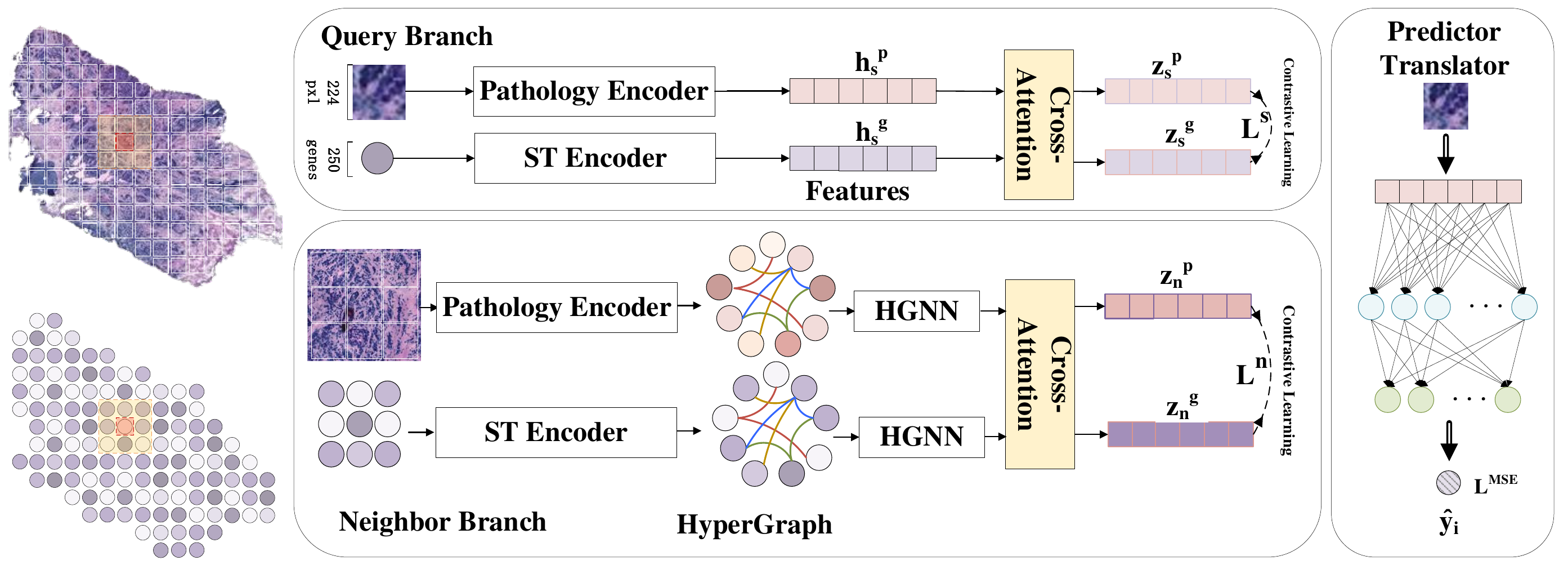}
    \caption{\textbf{The architecture of NH²2ST}. Our model consists of two key components: a query branch and a neighbor branch. It employs dual-level contrastive losses to strengthen the intrinsic connections between the two modalities, facilitating the final gene expression prediction. }
    \label{fig:framework}
\end{figure*}

\textbf{Framework.} NH²2ST consists of a query branch and a neighbor branch (see Fig. \ref{fig:framework}). The query branch extracts patch features \( h^p_s \in \mathbb{R}^{N} \) and gene features \( h^g_s \in \mathbb{R}^{N} \)  from pathology-gene pairs $\left<x_i,y_i \right>$ using pathology and ST encoders. These features are mutually refined via cross-attention and aligned through contrastive learning. In the neighbor branch, the patch \( x_i \) is considered alongside its \( K \) nearest neighboring patches and corresponding gene data. Two hypergraphs are constructed based on feature similarities and spatial proximity, and these graph representations are further refined and aligned through cross-attention and contrastive learning. Importantly, the patch features \( h^p_s \) in the query branch are further processed by a translator module to generate the final gene expression prediction \( \hat{y_i} \). Both branches share the pathology and ST encoders, with the neighbor branch serving as an auxiliary component. It enhances the pathology encoder while capturing relationships between multi-scale pathology and gene expression.

\textbf{Encoders.} Following previous studies \cite{TRIPLEX,BLEEP,yi2022approximate}, the pathology encoder $\phi_p : \mathbb{R}^{h \times w \times 3}$ $\to \mathbb{R}^{N}$ is a truncated pre-trained ResNet18 \cite{ResNet18}, extracting a \( 224 \times 224\) patch \( x_i \)  to generate an \( N \)-dimensional feature \( h^p \in \mathbb{R}^{N} \). The ST encoder \(\phi_g : \mathbb{R}^{n} \to \mathbb{R}^{N}\) consists of two fully connected layers, widely used to capture non-linear interactions, producing gene features \(h^g \in \mathbb{R}^{N} \).

\subsection{Query Branch}
This branch learns to predict gene expression values \( \hat{y} \) from the patch \( x \). 
It comprises a cross-attention module, cross-modal contrastive learning, and a prediction translator. 
Specifically, given the extracted feature pair $\left<h^p_s,h^g_s\right>$, the refined pathology features $z^{p}_{s}$ are computed as follows:  
\begin{equation}
    z^{p}_{s} = \phi^s_{ca}(h_s^p,h_s^g)= \text{softmax}\left(\frac{(\mathbf{W}^s_q  h^p_s)(\mathbf{W}^s_k h^g_s)^T}{\sqrt{d_k}}\right) (\mathbf{W}^s_v h_s^g),
\end{equation}
where the cross-attention function $\phi^s_{ca}(*,\circ)$ refines $*$ using guidance from $\circ$, $\mathbf{W}^s_q$, $\mathbf{W}^s_k$ and $\mathbf{W}^s_v$ are learnable projection matrices, \(d_k\) represents the dimensionality of the key vector projection $\mathbf{W}^s_kh^g_s$. Similarly, the refined gene features $z_s^g$ are obtained as $ z_s^g = \phi^s_{ca}(h^g_s, h^p_s)$. This bi-directional mechanism \cite{CA} enables each modality (\textit{e.g.}, pathology or gene) to attend to relevant patterns in the other, establishing semantic correspondences and aligning histological structures with transcriptomic profiles.

The refined feature pair $\left<z_s^p,z_s^g \right>$ is further aligned explicitly using cross-modal contrastive learning, where pairs from the same spatial location are treated as positive, while unpaired features serve as negative. The contrastive objective $\mathcal{L}^s$ is defined using the InfoNCE loss \cite{InfoNCE}:  \begin{equation}
    \mathcal{L}^s = \mathcal{L}(\left \{z_s^p\}^B,\{z_s^g\}^B \right) = -\frac{1}{B} \sum_{i=1}^{B} \log \frac{\mathcal{P}^{+}({z_s^p}_i, {z_s^g}_i)}{\mathcal{P}^{+}({z_s^p}_i, {z_s^g}_i) + \sum_{j=1}^{B-1} \mathcal{P}^{-}({z_s^p}_i, {z_s^g}_j)} ,
\end{equation}
where $B$ is the batch size, $\mathcal{P}^+(\cdot, \cdot) = \exp\left(\text{sim}(\cdot, \cdot) /\tau\right)$ denotes the similarity of positive pairs, $\mathcal{P}^-(\cdot, \cdot)  = \exp\left(\text{sim}(\cdot, \cdot) /\tau\right)$ represents the similarity of negative pairs, and $\tau$ is the temperature coefficient. Contrastive learning aligns pathology and gene features in a shared latent space while preserving biological relationships. It enhances discriminative representation by clustering same-spatial features while separating irrelevant ones, improving robustness to biological and technical noise (\textit{e.g.}, cell-type heterogeneity \cite{Cell-type-heterogeneity} and batch effects \cite{Batch-Size}).

Both cross-attention and contrastive learning are used to capture the intrinsic relationship between pathology and gene expression. Our goal is to predict gene expression using only pathology patches during inference. To achieve this, we introduce a prediction translator \(\phi_t: \mathbb{R}^{N} \rightarrow \mathbb{R}^{n}\) that converts the extracted pathology features \( h^p_s \) into predicted gene expression values $\hat{y} = \phi_t(h^p_s)$. The translator consists of two fully connected layers (opposite to the ST encoder), and is optimized using the Mean Squared Error (MSE) loss \cite{TRIPLEX,EGN}, denoted as: $ \mathcal{L}_{MSE} = \frac{1}{B} \sum_{i=1}^{B} (y_i - \hat{y_i})^2$.

In the training, the pathology encoder $\phi_p$ is simultaneously driven by the translator and contrastive learning. This creates a ``loop'' \cite{loop2}. Rather than causing instability, it typically enhances multimodal learning by enforcing representations that are both predictive (for the translator) and coherent across modalities (for contrastive alignment).

\subsection{Neighbor Branch}
This branch serves as an auxiliary component to assist the query branch in training pathology encoders $\phi_p$ by effectively utilizing neighbor information. It consists of hypergraph representation, a cross-attention module, and cross-modal contrastive learning. Given a patch \( x\), we select its \( K \) nearest spatial neighboring patches \( \{x_{j}\}_{j=1}^K \) and corresponding ST data. 
We then use a hypergraph to extract features from these neighboring patches. Unlike traditional graphs \cite{Hist2ST}, hypergraphs connect multiple nodes within a single edge, capturing higher-order relationships. This is inspired by groups of co-expressed genes extending beyond pairwise edges \cite{co-expression}, leading to a richer structure and more robust alignment.

Take pathology patches as an example, we construct a graph $\mathcal{G}^p = (\mathcal{V}, \mathcal{E})$, where $\mathcal{V} =\{v_1, \ldots, v_K\}$ represents the node set and $\mathcal{E} = \{e_1, \ldots, e_K\}$ denotes the hyperedge set. Each node $v_k$ corresponds to a patch $x_k$. Each hyperedge $e_j$ has a degree of $\tau$, connecting node $v_j$ with its $\tau-1$ most similar nodes. The similarity between nodes is calculated as follows:
\begin{equation}
    e_j = \{v_j\} \cup \left\{ v_k \ \middle| \ k \in \arg\max_{k \neq j}^{\tau-1} \text{sim}_{jk}(h_{nj}^p, h_{nk}^p) \right\} ,
\end{equation}
where $\text{sim}(\cdot,\cdot)$ represents the cosine similarity. Similarly, a gene hypergraph \( \mathcal{G}^g \) is constructed. Both hypergraphs are processed through separate \( L \)-layer HGNNs and a pooling layer, yielding paired graph representations $\left< h^p_n\in \mathbb{R}^N, h^g_n \in \mathbb{R}^N \right>$. The hypergraph models both the spatial morphology of pathology patches and gene co-expression patterns \cite{co-expression}, capturing relationships between a patch and its surrounding tissue structures, as well as between gene expression profiles in spatially adjacent regions.

Then, similar to the query branch, the cross-attention module \(\phi^n_{ca}\) is applied to obtain refined graph features $\left< z^p_n, z^g_n \right>$ as followed by:
\begin{equation}
z_n^p = \phi^n_{ca}(h^p_n, h^g_n),   z_n^g = \phi^n_{ca}(h^g_n, h^p_n).  
\end{equation}
By incorporating cross-attention at the neighbor level through hypergraph, we enrich contextual integration while capturing structural intricacies. This enhances the alignment of multimodal relationships. The refined features $\left< z^p_n, z^g_n \right>$ are then used for contrastive learning as: $\mathcal{L}^n = \mathcal{L}(\{z_n^p\}^B,\{z_n^g\}^B)$. By aligning the two modalities at the neighborhood level, it enhances the capture of broader spatial and co-expression interactions.

\subsection{Training and Inference}
The training objective consists of contrastive learning losses from both the query and neighbor branches, along with the translation loss in the query branch. The total loss $\mathcal{L}$ is computed as:
\begin{equation}
    \mathcal{L} =  \lambda_1 \cdot \mathcal{L}^{s} + \lambda_2 \cdot \mathcal{L}^{n} + \mathcal{L}_{MSE} ,
\end{equation}
where $\lambda_1$ and $\lambda_2$ are balancing weights. During inference, since gene expression data is unavailable, only the Predictor Translator is used to predict gene expression values $\hat{y_i} = \phi_g(\phi_p(x_i))$ based on input patch $x_i$.

\section{Experiments}

\begin{table*}[t]
    \caption{Comparison of our model with advanced models on six datasets.}
    \centering
    \resizebox{\textwidth}{!}{%
    \begin{tabular}{c|c|c|c|c|c|c}
\toprule
\midrule
\multicolumn{1}{c|}{\multirow{2}[0]{*}{Models}} & {MSE ($\downarrow$)} & {MAE ($\downarrow$)} & {PCC ($\uparrow$)} & {MSE ($\downarrow$)} & {MAE ($\downarrow$)} & {PCC ($\uparrow$)} \\
\cmidrule{2-7}
  & \multicolumn{3}{c|}{{STNet dataset}} & \multicolumn{3}{c}{{Skin}} \\
\cmidrule{1-7}
ST-Net~\cite{ST-Net} & 0.209±0.02 & 0.349±0.02 & 0.223±0.10 & 0.294±0.07 & 0.428±0.05 & 0.382±0.08 \\
EGN~\cite{EGN} & 0.192±0.02 & 0.337±0.04 & 0.203±0.09 & 0.281±0.08 & 0.418±0.06 & 0.388±0.06 \\
BLEEP~\cite{BLEEP} & 0.235±0.02 & 0.369±0.02 & 0.193±0.10 & 0.297±0.08 & 0.430±0.04 & 0.396±0.08 \\
HistoGene~\cite{HistToGene} & 0.199±0.03 & 0.335±0.04 & 0.201±0.09 & 0.294±0.07 & 0.415±0.07 & 0.406±0.10 \\
His2ST~\cite{Hist2ST} & 0.181±0.07 & 0.333±0.02 & 0.199±0.07 & 0.291±0.16 & 0.924±0.29 & 0.353±0.07 \\
TRIPLEX~\cite{TRIPLEX} & 0.202±0.07 & 0.343±0.02 & 0.352±0.10 & 0.268±0.09 & 0.404±0.07 & \textbf{0.490±0.07} \\
NH²2ST (our)& \textbf{0.161±0.03} & \textbf{0.311±0.02} & \textbf{0.572±0.10}   & \textbf{0.208±0.15} & \textbf{0.349±0.08} & 0.478±0.11 \\

\midrule
\end{tabular}}
\resizebox{\textwidth}{!}{%
\begin{tabular}{c|c|c|c|c|c|c}
\multicolumn{1}{c}{Models} & \multicolumn{3}{|c}{{INT(HEST-1k)}} & \multicolumn{3}{|c}{{ZEN(HEST-1k)}}\\
\cmidrule{1-7}
ST-Net~\cite{ST-Net} & \textbf{0.138±0.02} & \textbf{0.278±0.04} & 0.213±0.06 & \textbf{0.079±0.01} & \textbf{0.159±0.01} & 0.155 ± 0.01 \\
BLEEP~\cite{BLEEP} & 0.403±0.04 & 0.333±0.06 & 0.104±0.07 & 0.172±0.01 & 0.200±0.01 & 0.229±0.20 \\
TRIPLEX~\cite{TRIPLEX}  & 0.221±0.01 & 0.390±0.03 & 0.184±0.01& 0.099±0.02 & 0.150±0.01 & \textbf{0.242±0.03} \\
NH²2ST (our) & 0.220±0.03 & 0.388±0.03 & \textbf{0.368±0.02} & 0.108±0.01 & 0.193±0.01 & 0.216±0.02  \\
\midrule
\end{tabular}}

\resizebox{\textwidth}{!}{%
\begin{tabular}{c|c|c|c|c|c|c}
\multicolumn{1}{c}{{Models}} & \multicolumn{3}{|c}{{PCW(STimage-1K4M)}} & \multicolumn{3}{|c}{{Mouse(STimage-1K4M)}}\\
\cmidrule{1-7}
ST-Net~\cite{ST-Net} & 0.034±0.01 & 0.139±0.01 & 0.466±0.02 & 0.041±0.01 & 0.147±0.01 & 0.256±0.01\\
BLEEP~\cite{BLEEP} & 0.166±0.09 & 0.295±0.05 & 0.241±0.04 & 0.085±0.01 & 0.219±0.01 & 0.165±0.01 \\
TRIPLEX~\cite{TRIPLEX}   & 0.039±0.01 & 0.144±0.01 & 0.512±0.01 & 0.152±0.03 & 0.304±0.01 & 0.200±0.01 \\
NH²2ST (our) & \textbf{0.029±0.01} & \textbf{0.129±0.02} & \textbf{0.560±0.03} & \textbf{0.039±0.02} & \textbf{0.107±0.01} & \textbf{0.593±0.01}  \\
\midrule
\bottomrule
\end{tabular}}

\label{table: cross comparison}
\end{table*}

We conducted experiments on: ST-Net \cite{ST-Net}, Skin \cite{skin}, two datasets from STimage-1K4M \cite{STimage-1K4M}, and two datasets from HEST-1k \cite{HEST-1k}. We followed the previous data split settings \cite{TRIPLEX,HEST-1k}. Image patches were resized to $224\times224$ pixels, and 250 highly expressed genes were selected. Adam optimizer was employed with an initial learning rate of 0.0001, dynamically adjusted with a step size of 50 and a decay rate of 0.9. The batch size was 8 and the training epoch was 20.

We evaluated model performance using three metrics \cite{TRIPLEX}: Mean Squared Error (MSE) for error magnitude, Mean Absolute Error (MAE) for average absolute differences, and Pearson Correlation Coefficient (PCC) for linear correlation. Lower MSE and MAE indicate higher accuracy, while higher PCC shows stronger consistency between predictions and actual values. We conducted $k$-fold cross-validation and reported results as mean $\pm$ standard deviation (STD).

\subsection{Results and Discussion}
We compared our model against several advanced methods (see Table \ref{table: cross comparison}). Results for the ST-Net and Skin datasets were quoted from TRIPLEX \cite{TRIPLEX}, and results for other datasets were reproduced using the released codes.
Our model achieved the best performance across three metrics on STNet, Skin, and two STimage-1K4M datasets, and comparable results on HEST-1k. Notably, PCC results showed significant improvements over previous methods across all datasets. 
We attribute these gains to the neighbor branch with graph learning, which explicitly models relationships between the query image and adjacent regions, and the use of contrastive learning and cross-attention, which enhances interactions between pathology and gene modalities.

\subsection{Ablation Studies}

\begin{table*}[t]
    \centering
    \caption{\textbf{Left:} Ablation studies of branch construction. Variables tested include: query ($Q.$) and neighbor ($N.$) branch, cross-attention ($CA.$), contrastive learning ($CL.$), graph types including graph (G.) and hypergraph($HG.$), self-attention ($SA.$). \textbf{Right:} Results for different $K$ nearest patches and $L$-layer HGNN in hypergraph.}  
    \label{tab:abl}
    \small 
    
    \begin{minipage}[t]{0.52\linewidth} 
    \centering
    \resizebox{\linewidth}{!}{%
   \begin{tabular}{ ccc |c c c}
\toprule
\midrule
Branch & $CA.$ & $CL.$ & MSE ($\downarrow$) & MAE ($\downarrow$) & PCC ($\uparrow$)  \\
\midrule
$Q.$ & & & 0.147±0.06 & 0.331±0.06 & 0.320±0.22  \\
$Q.$ & $SA.$ &  & 0.087±0.01 & 0.224±0.02 & 0.155±0.01   \\
$Q.$ & & \checkmark & 0.781±0.06 & 0.809±0.11 & 0.424±0.03  \\
$Q.$ & \checkmark & \checkmark & 0.093±0.04 & 0.229±0.06 & 0.329±0.06  \\
\cmidrule(lr){1-6}
$Q.+N.$ & \checkmark & & 0.106±0.07 & 0.195±0.05 & 0.468±0.05\\
$Q.+N.$ &  & \checkmark & 0.597±0.05 & 0.703±0.04 & 0.451±0.03 \\
$Q.+N.$ ($G.$) & \checkmark & \checkmark & 0.043±0.02 & 0.154±0.06 & 0.411±0.02 \\
$Q.+N.$ ($HG.$) & $SA.$ & \checkmark & 0.035±0.01 & 0.143±0.01 & 0.526±0.03 \\
\cmidrule(lr){1-6}
\multicolumn{3}{c|}{NH²2ST} & \textbf{0.029±0.01} & \textbf{0.129±0.02} & \textbf{0.560±0.03}  \\ 
\midrule
\bottomrule
\end{tabular}
    }
    \end{minipage}
    \hfill
    \begin{minipage}[t]{0.46\linewidth}
    \centering
    \resizebox{\linewidth}{!}{
   \begin{tabular}{ c|c c c}
\toprule
\midrule
Configuration & MSE ($\downarrow$) & MAE ($\downarrow$) & PCC ($\uparrow$) \\
\cmidrule(lr){1-4}
 $K$=4 & 0.034±0.01 & 0.142±0.01 & 0.525±0.03 \\
$K$=8 & \textbf{0.033±0.01} & \textbf{0.141±0.01} & \textbf{0.530±0.03} \\
$K$=16 & 0.035±0.01 & 0.144±0.01 & 0.529±0.02 \\
$K$=25 & 0.036±0.01 & 0.146±0.01 & 0.526±0.03 \\
\cmidrule(lr){1-4}
$L$=1 & 0.062±0.01 & 0.191±0.02 & 0.469±0.04 \\
$L$=2 & \textbf{0.033±0.01} & \textbf{0.141±0.01} & \textbf{0.530±0.02} \\
$L$=3 & 0.093±0.04 & 0.231±0.05 & 0.256±0.14 \\
$L$=4 & 0.130±0.05 & 0.276±0.06 & 0.207±0.09 \\
\cmidrule(lr){1-4}
NH²2ST & 0.029±0.01 & 0.129±0.02 & 0.560±0.03 \\ 
\midrule
\bottomrule
\end{tabular}
    }
    \end{minipage}

\end{table*}


\textbf{Branch Construction.}
We evaluated the impact of different branch components on PCW dataset (see Table \ref{tab:abl}). When using only the query branch, incorporating self-attention ($SA.$)  improved MSE and MAE but reduced PCC by 0.165, while using contrastive learning ($CL.$) increased PCC by 0.104 but significantly raised errors. Integrating $CL.$ with cross-attention $CA.$ improved all three metrics. For the neighbor branch, a similar trend was observed, using only $CA.+CL.$ led to significant performance gains. Additionally, we examined the effect of graph ($G.$) or hypergraphs ($HG.$) with self-attention in the neighbor branch. While both performed reasonably well, they were inferior to our proposed method. These results highlight that the combination of contrastive learning and cross-attention is crucial, as cross-attention facilitates interactions between modalities.

Moreover, we tested different neighbor sizes and HGNN layers, observing the best performance at \(K=8\) and \(L=2\). We attribute this to the fact that excessive neighbors introduce too much regional information, while more HGNN layers lead to over-smoothing of node information.

\noindent \textbf{Contrastive Learning.}
We explored the impact of different contrastive learning 
 on the PCW dataset (see Table \ref{table: Contrastive Learning}). We observed that all three metrics achieved optimal results with a batch size of 8, particularly PCC, which reached 0.550. As the batch size increased further, no significant improvements were observed. Regarding the balance parameters, setting $\lambda_1=1$ and $\lambda_2=0.5$ yielded the best results. While $\lambda_1=0$ and $\lambda_2=1$ slightly improved MSE and MAE, it led to 7.4\% lower PCC compared to our chosen configuration. This indicates a crucial trade-off between the two loss functions in determining the final performance. For the temperature coefficient, a value of 0.05 yielded the best performance. Higher values resulted in slightly reduced performance but remained relatively stable.

\begin{table*}[t]
\centering
\caption{\textbf{Ablation study of contrastive learning}. Investigating the impact of batch size ($B$), loss balancing hyperparameters $(\lambda_1, \lambda_2)$, and temperature coefficient $\tau$.}
\resizebox{\textwidth}{!}{%
\begin{tabular}{c|c c c|c|c c c|c|c c c}
\toprule
\midrule
 
$B$ & MSE ($\downarrow$) & MAE ($\downarrow$) & PCC ($\uparrow$) & ($\lambda1$,$\lambda2$) & MSE ($\downarrow$) & MAE ($\downarrow$) & PCC ($\uparrow$) & $\tau$ & MSE ($\downarrow$) & MAE ($\downarrow$) & PCC ($\uparrow$)\\
\cmidrule(lr){1-12}
4 & 0.182±0.03 & 0.300±0.02 & 0.397±0.01 &(0,1) & \textbf{0.042±0.01} & \textbf{0.150±0.01} & 0.470±0.04  & 0.025  & 0.191±0.03 & 0.333±0.02
& 0.228±0.19\\
8 & \textbf{0.030±0.01} & \textbf{0.132±0.01} & \textbf{0.550±0.02}  & (1,0)  & 0.093±0.04 & 0.229±0.05 & 0.353±0.08  & 0.05 & \textbf{0.039±0.01} & \textbf{0.152±0.01} & \textbf{0.505±0.04}   \\
16 & 0.034±0.01 & 0.144±0.01 & 0.534±0.02  & (0.5,1)  & 0.050±0.01 & 0.173±0.010 & 0.494±0.02  & 0.1 & 0.062±0.01 & 0.191±0.02 & 0.469±0.04  \\
32 & 0.062±0.01 & 0.191±0.02 & 0.469±0.03 & (1,0.5)  & 0.043±0.01 & 0.160±0.01 & \textbf{0.508±0.03}
 & 0.15 & 0.043±0.01 & 0.158±0.01 & 0.483±0.03  \\
64 & 0.068±0.01 & 0.284±0.09 & 0.482±0.06  & (1,1) & 0.061±0.01 & 0.193±0.02 & 0.469±0.04 &0.2 & 0.039±0.01 & 0.153±0.01 & 0.491±0.02\\
\cmidrule(lr){1-12}
 NH²2ST & 0.029±0.01 & 0.129±0.02 & 0.560±0.03 & NH²2ST & 0.029±0.01 & 0.129±0.02 & 0.560±0.03 & NH²2ST & 0.029±0.01 & 0.129±0.02 & 0.560±0.03\\ 
\cmidrule(lr){1-12}
\bottomrule
\end{tabular}}
\label{table: Contrastive Learning}
\end{table*}

\begin{figure*}[t]
    \centering
    \includegraphics[width=1\textwidth]{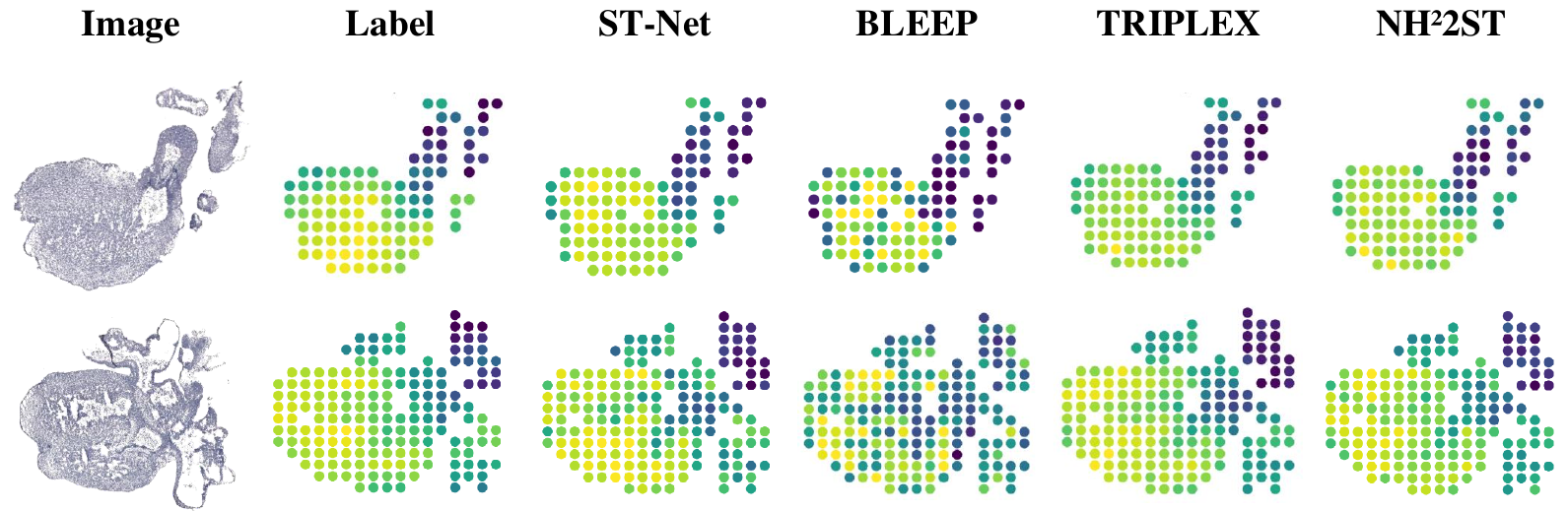}
    \caption{\textbf{Visualization of expression values on the PCW Dataset}. From left to right: raw WSI, labels, ST-Net, BLEEP, TRIPLEX and our NH²2ST. }
    \label{fig:visualization}
\end{figure*}

\noindent \textbf{Visualization.} We conducted a visualization comparison of our model against other methods on the PCW dataset (see Fig. \ref{fig:visualization}). Compared to BLEEP (contrastive learning-based) and TRIPLEX (neighbor-based), our model exhibited higher visual consistency with the true values, particularly in highly expressed regions, demonstrating its effectiveness in accurate prediction.

\section{Conclusion}

We present NH²2ST that integrates dual-scale spatial modeling and contrastive learning to predict spatially resolved gene expression from whole slide images. It leverages hypergraphs to capture contextual relationships among adjacent regions and employs cross-attention contrastive learning to refine and align pathology and gene expression modalities. Experiments on six publicly available datasets demonstrate the superior performance of our framework.

\textbf{Limitation.}
 NH²2ST utilizes only the query branch during inference, as experiments show that incorporating the neighborhood branch may introduce complexity or noise, leading to performance degradation. Additionally, the framework relies on well-aligned pathology images and ST data, with misalignment or noise could impact accuracy. Future work may explore effective integration of neighborhood information during inference and develop robust preprocessing techniques to mitigate data misalignment and noise.

\bibliographystyle{splncs04}
\bibliography{mybib}

\end{document}